\documentclass[runningheads]{llncs}
\usepackage[T1]{fontenc}
\usepackage{graphicx}
\usepackage{hyperref}
\usepackage{adjustbox}
\usepackage{algorithm}
\usepackage{algpseudocode}
\usepackage{aliascnt}
\usepackage{amsfonts}
\usepackage{amsmath}
\usepackage{multirow}
\usepackage{multicol}
\usepackage{tabularx}
\usepackage{xcolor}
\usepackage{amsmath}

\begin{document}
\title{Aspect-Aware MOOC Recommendation in a Heterogeneous Network}
%
%
\author{Seongyeub Chu, Jongwoo Kim, \and Mun Yong Yi\thanks{Corresponding author.}}
\institute{Korea Advanced Institute of Science and Technology\\
\email{\{chseye7, gsds4885, munyi\}@kaist.ac.kr}}
%
%
\maketitle              
\begin{abstract}
MOOC recommendation systems have received increasing attention to help learners navigate and select preferred learning content. Traditional methods such as collaborative filtering and content-based filtering suffer from data sparsity and over-specialization. To alleviate these limitations, graph-based approaches have been proposed; however, they still rely heavily on manually predefined metapaths, which often capture only superficial structural relationships and impose substantial burdens on domain experts as well as significant engineering costs. To overcome these limitations, we propose AMR (Aspect-aware MOOC Recommendation), a novel framework that models path-specific multiple aspects by embedding the semantic content of nodes within each metapath. AMR automatically discovers metapaths through bi-directional walks, derives aspect-aware path representations using a bi-LSTM–based encoder, and incorporates these representations as edge features in the learner-learner and KC-KC subgraphs to achieve fine-grained semantically informed KC recommendations. Extensive experiments on the large-scale MOOCCube and PEEK datasets show that AMR consistently outperforms state-of-the-art graph neural network baselines across key metrics such as HR@K and nDCG@K. Further analysis confirms that AMR effectively captures rich path-specific aspect information, allowing more accurate recommendations than those methods that rely solely on predefined metapaths. The code will be available upon accepted.

\keywords{MOOC Recommendation  \and Graph Neural Network \and Aspect \and Heterogeneous Network}
\end{abstract}
\section{Introduction}
\label{intro}
The rise of massive open online courses (MOOCs) is transforming global education, with platforms like EdX\footnote{\url{https://www.edx.org/}} and Coursera\footnote{\url{https://www.coursera.org/}} offering diverse courses across various fields. While this variety is beneficial, it also creates challenges for learners in selecting courses that align with their educational and professional goals. As a result, personalized recommendation systems are gaining attention to help guide learners to courses suited to their individual interests and objectives.

In personalized recommendation systems, it’s essential to suggest course content that aligns with learners' objectives and is likely to be completed successfully. To this end, numerous researchers have explored graph-based approaches to alleviate data sparsity by modeling relationships between entities \cite{piao,ju,gong,yu}. MOOC data, involving learners, videos, knowledge concepts (KCs), and courses, is often represented as a heterogeneous graph, with metapaths used to capture meaningful relational patterns. However, constructing effective metapaths typically requires hand-crafted design or extensive domain-specific expertise \cite{kim2024leveraging}, which places a substantial burden on experts. Furthermore, previous studies using this paradigm have been limited in that they primarily capture superficial structural connections derived only from node types, while overlooking the distinct semantic characteristics contained within nodes along each metapath.

This study proposes AMR (Aspect-aware MOOC Recommendation), a framework that uses multiple path-specific aspects to improve the accuracy of the recommendation in heterogeneous networks and overcome the limitations of traditional metapath-based methods. An aspect is defined as a vector representation of a metapath encoding its node contents, with the number of aspects predefined to distinguish embeddings for identical node types based on their content. Our goal is to recommend knowledge concepts (KCs) by estimating learner-provided ratings, as directly suggesting courses or videos often fails to address the specific conceptual needs of the learners \cite{piao,gong}. AMR captures aspect information from metapaths automatically discovered via bi-directional walks, integrates them as edge features in learner-learner and KC-KC subgraphs, and uses a bi-directional LSTM to aggregate node embeddings into aspect-based path representations. These representations are fused by an aspect aggregator to enrich the subgraph features, enabling accurate and semantically informed KC recommendations. 

We conducted experiments on two widely used real-world MOOC benchmarks, MOOCCube \cite{yu2020mooccube} and PEEK \cite{bulathwela2021peek}, comparing AMR against multiple baselines. Our findings show that AMR outperforms previous methods that leverage graph neural network approaches. Our contributions are as follows: (1) We introduce a bi-directional walk which automatically identifies metapaths connecting nodes of the same type (either learner-learner or KC-KC). (2) We propose AMR, which leverages path-specific aspects to provide more fine-grained and accurate KC recommendation. (3) We conduct extensive experiments on the MOOCCube and PEEK datasets. The results show that AMR not only exceeds strong baselines, but also captures diverse aspects along the paths that connect the KCs, leading to richer and more discriminative representations.


    


\section{Related Work}
\label{related_work}

\subsection{Graph-based Recommendation System on MOOC Platform}
Graph-based recommendation models have become prevalent in the MOOC domain to address the limitations of traditional content-based and collaborative filtering approaches \cite{magron2022neural,bourgais2022avoiding,tian2024content}. MOOC data is represented as a heterogeneous information networks (HINs), with graph neural networks (GNN) used to capture learner and KC representations through metapaths \cite{gong,piao,ju}. Attention mechanisms aggregate information from metapaths to refine learner-KC relations \cite{gong}, while latent learner and KC information is incorporated into attention outputs in another study \cite{piao}. Additionally, semantic entity relationships are sampled to enrich node information, and the heterogeneous graph is partitioned into metapath-based subgraphs, integrating learner and KC embeddings via attention \cite{ju}. A recent study \cite{wang2022attentional} further introduces structural neighbor–enriched contrastive learning across metapaths and employs InfoNCE loss to reduce learner-interest bias. Despite these advances, most MOOC studies rely on predefined metapaths that require substantial domain expertise, which may overlook rich, unanticipated relationships and limit the ability to capture meaningful differences between entities, reducing recommendation effectiveness.


\subsection{Multi-aspect Network Embedding}

In HINs, objects can exhibit multiple aspects, and different relationships can highlight distinct aspects with varying semantics. Several studies have incorporated multi-aspect modeling into graph embeddings. For instance, ASPEM \cite{shi2018aspem} embeds each aspect of a node separately to address semantic incompatibility between aspects, while MCNE \cite{wang2019mcne} uses a binary mask layer to decompose a vector into multiple conditional embeddings and applies attention to capture inter-aspect interactions. Other approaches assign an embedding vector to each node facet \cite{liu2019single} or dynamically infer node aspects from local context using an end-to-end framework based on Gumbel–Softmax \cite{park2020unsupervised}. However, most MOOC-related studies using HINs rely on predefined metapaths that require extensive domain expertise, imposing substantial burdens on experts, and overlook fine-grained node-level information along intermediate paths, ultimately leading to suboptimal performance. To overcome this, we propose a new approach that explores multiple aspects of diverse metapaths within HINs without relying on fixed, manually defined metapaths.

\section{Problem Statement}
\label{problem_statement}
For a specific learner $l$ with associated interactive data in MOOCs, 
the objective is to estimate the learner's interest score over a set of KCs 
and subsequently generate a top-$N$ list of recommended KCs. 
Formally, given the interactive data of the learner $l$, we train a prediction function, $f(l) \rightarrow \{ k_i \mid k_i \in \mathcal{K},\, i \le N \}$, where $\mathcal{K}$ represents the pool of candidate KCs (e.g., ``database,'' ``algorithms''). Applying $f$ to the learner $l$ yields an individualized recommendation list of KCs ranked according to predicted interest scores.



\section{Aspect-aware MOOC Recommendation (AMR)}
\label{proposed_method}

\begin{figure*}[t]
\centerline{\includegraphics[width=\textwidth ]{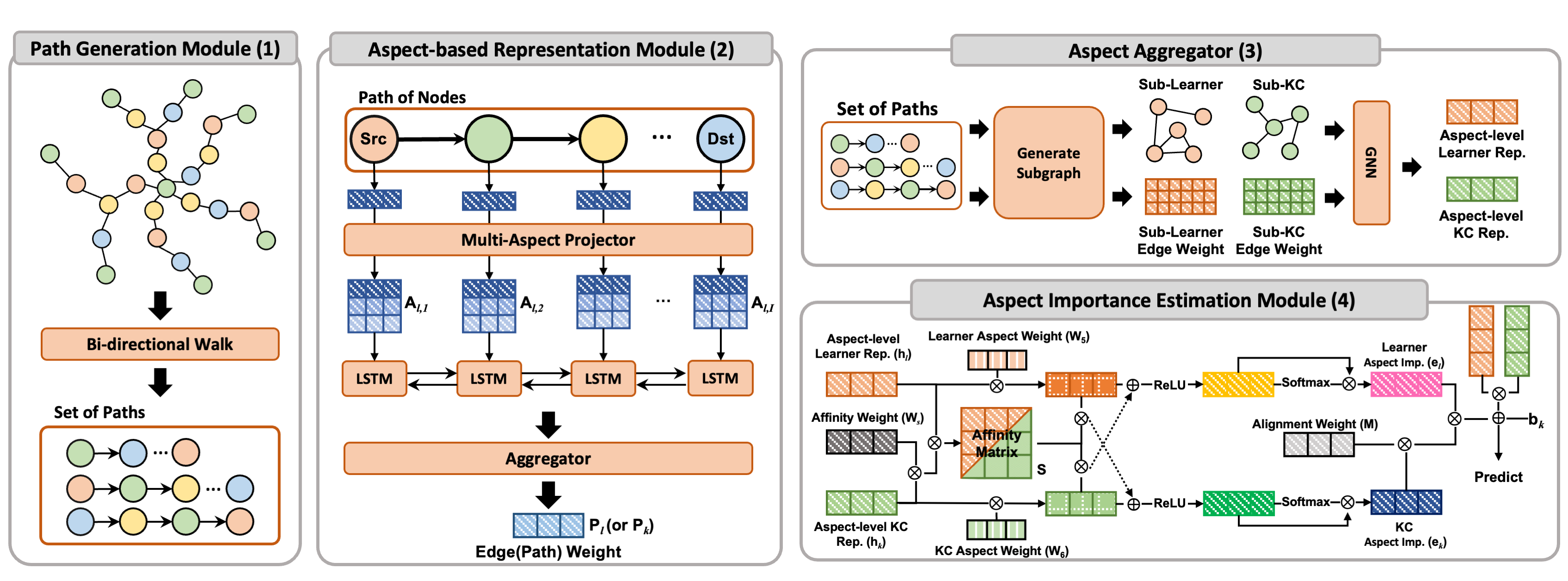}}
\caption{Overall Architecture of AMR. The framework consists of four key components which are (1) path generation module, (2) aspect-based representation module, (3) aspect aggregator, and (4) aspect importance estimation module.}
\label{fig2}
\end{figure*}

The architecture of our proposed KC recommendation system, AMR, is shown in Fig.~\ref{fig2}. AMR utilizes relational and aspect information in MOOC networks, including learners and KCs. The framework consists of four key components:
(1) a path generation module that extracts metapaths using bi-directional walks;
(2) an aspect-based representation module that derives aspect-specific representations of learners and KCs;
(3) an aspect aggregator that integrates representations via a GNN;
(4) an aspect importance estimation module that models learner-KC interactions to estimate aspect importance.

\subsection{Feature Extractor} \label{feature_extractor}
The names of KCs from MOOCs contain semantic information, so we extract word embeddings using FastText \cite{bojanowski2017enriching} and use them as content features for KCs. We then capture structural characteristics by analyzing the relationships among entities in the MOOC dataset (e.g., learners, KCs, videos). This allows us to extract relational patterns that complement the semantic representations of KCs and construct a heterogeneous graph with distinct node types for each entity. We derive the features of non-KC entities through their interactions with KCs in the graph, projecting KC content features into a common latent space using adjacency matrices. Specifically, we compute $X_A = R_{A,K} X_K^{\top}$, where $X_A$ is the projected feature matrix of entity $A$, $R_{A,K}$ is the adjacency matrix between entity A and KCs, and $X_K$ is the KC content feature matrix.

\subsection{Bi-directional Walk}

We introduce a bi-directional walk algorithm, which automatically identifies metapaths connecting nodes of the same type (either learner–learner or KC–KC). It uncovers connections where learners or KCs are linked via intermediate nodes. The method operates by performing repeated inner products from both ends: one starting from a node of a given type (e.g., learner) and the other from another node of the same type. These expansions explore neighboring nodes of different types and gradually construct candidate paths. The process continues until the two expansions meet at an intermediate node, forming a complete path. This simultaneous exploration from both directions is why we call it the bi-directional walk. To enumerate and store paths, we apply a breadth-first search (BFS) on the processed adjacency matrices from the iterative inner products, generating as many paths as specified by the hyperparameter~$p$ (set to 10 in this study).

\subsection{Aspect-based Representation Module} \label{aspect_representation}
Not all intermediary nodes in a path are equally significant; their importance varies based on the aspect being considered. Often, entities connecting learners or KCs relate to multiple aspects of the target nodes. Therefore, focusing on the aspects most relevant to the endpoint nodes yields aspect-level representations that capture specific information from the intermediary nodes. Motivated by previous work in the recommendation domain that demonstrated strong performance using aspect-based representations for reviews \cite{Cheng18a,Cheng18b,kim2024leveraging}, we hypothesize that the incorporation of aspects can further enhance the effectiveness of KC recommendations. 

In what follows, we describe how the aspect-based path representations
$\mathbf{P}_{l}$ and $\mathbf{P}_{k}$ are obtained for a learner $l$ and a KC $k$. Because the procedures for learners and KCs are identical, we focus on the representation of learners. 
Each intermediate node on a path is associated with a $d$-dimensional embedding that is shared across all aspects ($a \in A$).
To allow for aspect-specific variations, we introduce an aspect-specific node projection matrix 
$\mathbf{W_{a}} \in \mathbb{R}^{d \times h}$, 
where $h$ is a hyperparameter indicating the number of latent factors for aspect-level representations, 
independent of the original node embedding dimension $d$. 
The transformation is defined as $\mathbf{A}_{l,i} = \mathbf{E}_{l,i}\,\mathbf{W}_{a}$, where $\mathbf{E}_{l,i} \in \mathbb{R}^{1 \times A \times d}$ denotes the initial $d$-dimensional embedding of the $i$-th neighbor node ($i \in I$) of the learner $l$ across all $A$ aspects
and $\mathbf{A}_{l,i} \in \mathbb{R}^{1 \times A \times h}$ represents the aspect-based integrated embedding that combines the information from all aspects of that node. Hereafter, the integrated embedding ($\mathbf{A}_{l,i}$) is referred to as an aspect-based embedding.

After obtaining the aspect-based embeddings of the nodes within a path between learner nodes, 
we feed them into bidirectional long short-term memory (bi-LSTM) layers. These layers process the embeddings along the path to produce representations of the target nodes and the nodes forming the paths leading to them. As a result, we obtain a latent aspect-based path representation vector
$\mathbf{\tilde{P}}_{l} \in \mathbb{R}^{I \times A \times h}$ for a learner $l$ with $I$ neighboring nodes. A latent aspect-based path representation vector $\mathbf{\tilde{P}}_{k}$ for a KC is derived similarly. We reshape $\mathbf{\tilde{P}}_l$ and $\mathbf{\tilde{P}}_k$ from $I \times A \times h$ to $I \times (A \times h)$, where $I$ denotes the number of nodes forming the path, $A$ is the number of aspects, and $h$ is the vector dimension. Subsequently, the latent aspect-based path representation vectors of the learner ($\mathbf{\tilde{P}}_{l}$) and the KC ($\mathbf{\tilde{P}}_k$) are aggregated into a vector by weighting and combining the embeddings at nodes $I$ according to their path-level importance using an attention mechanism with weight matrix $\mathbf{W}_{\beta}$. The process for a learner ($l$) is as follows: 

{\footnotesize
\setlength{\abovedisplayskip}{2pt}
 \setlength{\belowdisplayskip}{2pt}
\begin{equation}
\begin{split}
\tilde{\mathbf{P}}_{l} =
    \operatorname{bi\text{-}LSTM}\!\left(\{\mathbf{A}_{l,i}\}_{i=1}^{I}\right),
\quad
\boldsymbol{\beta}_l =
\operatorname{softmax}\!\big(\mathbf{W}_\beta\,\tilde{\mathbf{P}}_{l}\big), 
\quad
\mathbf{P}_{l} =
\boldsymbol{\beta}_{l}^\top \tilde{\mathbf{P}}_{l},
\end{split}
\label{eq:aspect_path}
\end{equation}
}
where $\tilde{\mathbf{P}}_{l}\in\mathbb{R}^{I \times (A \times h)}$ is the latent aspect-based path representation of learner $l$ after transformation by the bi-LSTM layers;  
$\boldsymbol{\beta}_{l}\in\mathbb{R}^{I \times 1}$ denotes the attention weights (importance scores) of each aspect within the paths;  
and $\mathbf{P}_{l}\in\mathbb{R}^{1 \times (A \times h)}$ indicates the final aspect-based path representation, which serves as the edge weight between pairs of learner nodes. The final aspect-based path representation in terms of a KC $\mathbf{P}_{k}\in\mathbb{R}^{1 \times (A \times h)}$ is obtained in the same manner.


\subsection{Aspect Aggregator} \label{aspect_aggregator}
The aspect-based path representation captures interactions between a learner $l_{a}$ and its neighbor $l_{b}$ using the aspect-based representation module. This path representation serves as the edge weight, reflecting the relationship strength between learners across the network. We then construct a homogeneous graph in which learner nodes are connected by edges weighted using the aspect-based path representations aggregated from intermediary nodes along the metapaths. Using this refined network, we apply a GNN to learn discriminative learner representations, as it is well-suited for modeling complex node relationships and predicting individual node characteristics \cite{kipf,velickovic2017graph,hamilton2017inductive}. We use GCN architecture \cite{kipf} as our backbone model. Given a graph $G$ with node features $\mathbf{M}_{l,i}$ for each neighboring learner node $i \in \mathcal{N}(l)$ of a target learner node $l$ and edge weights $\mathbf{P}_{l,i}$ for the relationships between the target learner node and its neighboring learner nodes, the forward propagation of a GCN layer~\cite{kipf} with weight matrix $\mathbf{W}_{k}$ is formulated as follows:
{\footnotesize
\setlength{\abovedisplayskip}{2pt}
 \setlength{\belowdisplayskip}{2pt}
\begin{equation}
\begin{split}
    \mathbf{h}^0_l = \mathbf{M}_{l},
    \quad
    \mathbf{h}^k_l = \operatorname{ReLU}\!\left(
        \mathbf{W}_k \cdot \sum_{i \in \mathcal{N}(l)} \mathbf{h}^{k-1}_i \odot \mathbf{P}_{l,i}
    \right),
    \quad
    \mathbf{h}_l = \mathbf{h}^k_l,
\end{split}
\label{eq10}
\end{equation}
}
where 
$\mathbf{M}_l = \{\mathbf{M}_{l,i}\mid i\in\mathcal{N}(l)\}$  
is the set of initial node embeddings of all neighbors $i$ of the learner $l$, $\mathbf{h}_l\in\mathbb{R}^{A \times h}$ denotes an aspect-based learner representation projected by the GCN and $k$ denotes the number of GCN layers. The same procedure is applied to construct the aspect-based representation for a KC.

\subsection{Aspect Importance Estimation Module} \label{aspect_importance_estimation}

This module jointly learns the aspect importance of learners and KCs. The aspect-based representations of KCs help determine the significance of each aspect to a learner, while the aspect-based representations of learners inform the significance of each aspect to a KC. The output is an $A$-dimensional vector representing the relevance of each aspect to a learner, and a corresponding $A$-dimensional vector indicating the relevance of each aspect to a KC. 

To integrate the aspect-based representations of KCs when determining the importance of each aspect to a learner
(and conversely for KCs), it is necessary to capture how the target learner and KC align at the aspect level. 
Given the aspect-based learner representation $\mathbf{h}_{l} \in \mathbb{R}^{A \times h}$ and the KC representation $\mathbf{h}_{k} \in \mathbb{R}^{A \times h}$, 
we first compute an aspect-level affinity matrix, $\mathbf{S} = \operatorname{ReLU}\!\left(\mathbf{h}_{l}\,\mathbf{W}_{s}\,\mathbf{h}_{k}^{\top}\right)$,
where $\mathbf{W}_{s} \in \mathbb{R}^{h \times h}$ is a learnable weight matrix
and the affinity matrix $\mathbf{S} \in \mathbb{R}^{A \times A}$ measures the mutual similarity
between each pair of aspect-based representations of learners and KCs.

Following prior work \cite{lu}, the affinity matrix $\mathbf{S}$ is then used as an additional feature to estimate the importance of the aspect in terms of the learners and the KCs, as shown in \ref{eq8}:
{\footnotesize
\setlength{\abovedisplayskip}{2pt}
 \setlength{\belowdisplayskip}{2pt}
\begin{equation}
\begin{split}
    \mathbf{\tilde{h}}_{l} &= \operatorname{ReLU}\!\left(\mathbf{h}_{l}\,\mathbf{W}_{5} 
          + \mathbf{S}^{\top}\big(\mathbf{h}_{k}\,\mathbf{W}_{6}\big)\right),
    \quad
    \boldsymbol{\beta}_{l} = \operatorname{softmax}\!\left(\mathbf{\tilde{h}}_{l}\,\mathbf{v}_{1}\right),
    \quad
    \mathbf{e}_{l} = \mathbf{\tilde{h}}_{l}\boldsymbol{\beta}_{l},
    \\
    \mathbf{\tilde{h}}_{k} &= \operatorname{ReLU}\!\left(\mathbf{h}_{k}\,\mathbf{W}_{6} 
          + \mathbf{S}^{\top}\big(\mathbf{h}_{l}\,\mathbf{W}_{5}\big)\right),
    \quad
    \boldsymbol{\beta}_{k} = \operatorname{softmax}\!\left(\mathbf{\tilde{h}}_{k}\,\mathbf{v}_{2}\right),
    \quad
    \mathbf{e}_{k} = \mathbf{\tilde{h}}_{k}\boldsymbol{\beta}_{k},
\end{split}
\label{eq8}
\end{equation}
}
where $\mathbf{h}_{l}, \mathbf{h}_{k} \in \mathbb{R}^{A \times h}$ are the learner and KC aspect-based representation obtained from Section \ref{aspect_aggregator}. 
The matrices $\mathbf{W}_{5}, \mathbf{W}_{6} \in \mathbb{R}^{h \times m}$ and vectors $\mathbf{v}_{1}, \mathbf{v}_{2} \in \mathbb{R}^{m \times 1}$ are learnable weight parameters. Vectors $\boldsymbol{\beta}_{l}, \boldsymbol{\beta}_{k} \in \mathbb{R}^{A \times 1}$ represent the estimated aspect-importance weights in aspects $A$ for the learner $l$ and the KC $k$, respectively. Consequently, $\mathbf{e}_{l}, \mathbf{e}_{k} \in \mathbb{R}^{A \times m}$ denote the final aspect-level representations of the learner and the KC, respectively, incorporating aspect importance via element-wise multiplication of aspect-based representations and aspect importance weights.

\subsection{Rating Prediction Module} \label{rating_prediction}
Given the aspect-level representations $\mathbf{e}_{l}, \mathbf{e}_{k} \in \mathbb{R}^{A \times m}$ of a learner $l$ and a KC $k$, which incorporate aspect importance as described in section \ref{aspect_importance_estimation}, the learner's preference score ($\hat{\mathbf{y}}_{l,k}$) for the KC is predicted using $\mathbf{h}_{l}^{\top}\mathbf{h}_{k}
      + \mathbf{e}_{l}\mathbf{M}\,\mathbf{e}_{k}^{\top}
      + b_{k}$.  
The matrices $\mathbf{h}_{l}, \mathbf{h}_{k} \in \mathbb{R}^{A \times h}$ are the aspect-based representations of the learner and the KC, respectively, obtained from Section \ref{aspect_aggregator} before importance incorporation and used for matrix factorization, and $b_{k}$ is a bias term.  
The matrix $\mathbf{M}\in \mathbb{R}^{m \times m}$ is a learnable parameter that maps $\mathbf{e}_{l}$ into the same latent space as $\mathbf{e}_{k}$.

\subsection{Loss function}


We combine both triplet loss ($L_{triplet}$) and Bayesian personalized ranking (BPR) loss ($L_{bpr}$) to effectively distinguish between positive and negative samples during training. The combined loss ($L$) is given by $L = L_{bpr} + L_{triplet}$. Specifically, the triplet loss is calculated as $\sum_{(l, p, n) \in D_S} (dist(\mathbf{h}_l, \mathbf{h}_p) - dist(\mathbf{h}_l, \mathbf{h}_n) + 1)$, where $D_S$ is the entire dataset, $dist$ denotes Euclidean distance, $\mathbf{h}_l$ is the learner's aspect-based representation, and $\mathbf{h}_p$ and $\mathbf{h}_n$ are the aspect-based representations of the positive and negative KC samples, respectively. The BPR loss ($L_{bpr}$) is computed as $-\sum_{(l, p, n) \in D_S} \log \sigma (\mathbf{\hat{y}}_{l,p} - \mathbf{\hat{y}}_{l,n})$, where $\hat{\mathbf{y}}_{l,p}$ and $\hat{\mathbf{y}}_{l,n}$ are the predicted preference scores for the positive and negative KC samples in $D_S$.

\section{Experiment}
\label{experiment}
We conduct extensive experiments to explore the multiple aspects embedded in metapaths and evaluate their effectiveness in MOOC recommendation. The investigation addresses following research questions:

\begin{itemize}
    \item \textbf{RQ1.} To what extent does leveraging multiple aspects of learners-KC relationships improve the performance of the MOOC recommendation?
    \item \textbf{RQ2.} What key insights can be drawn from analyzing the multiple aspects extracted from the metapaths connecting learners and KCs?
\end{itemize}

\subsection{Dataset}
\begin{table}[t]
\centering
\scriptsize
\caption{Statistics of entities and relations for MOOCCube and PEEK Dataset}
\begin{tabular}{|c|c|c|c|c|}
\hline
\textbf{Dataset} & \textbf{Entities} & \textbf{Statistics} & \textbf{Relations} & \textbf{Statistics} \\ \hline
         & learner           & 2,005  & learner-course            & 13,696  \\
         & video             & 22,403 & course-video              & 42,117  \\
MOOCCube & course            & 600    & teacher-course            & 1,875   \\
         & teacher           & 1,385  & video-knowledge concept   & 295,475 \\
         & KC & 21,037 & course-knowledge concept  & 150,811 \\ \hline
         & learner           & 4,063  & learner-lecture           & 101,741 \\
PEEK     & lecture           & 23,200 & lecture-knowledge concept & 470,466 \\
         & KC & 23,200 &                           &         \\ \hline
\end{tabular}%
\label{table_1}
\end{table}

We use two real-world MOOC benchmark datasets: MOOCCube \cite{yu2020mooccube} and PEEK \cite{bulathwela2021peek}. MOOCCube, collected from the XuetangX platform\footnote{https://www.xuetangx.com/}, contains rich data on learner activities and MOOC content, including five entity types—\textit{learner}, \textit{video}, \textit{course}, \textit{teacher}, and \textit{KC}—modeled as a heterogeneous graph. Each course or video is linked to KCs. PEEK, collected from VideoLectures.Net\footnote{www.videolectures.net
}, includes three entity types—\textit{learner}, \textit{lecture}, and \textit{KC}—also modeled as a heterogeneous graph. Dataset statistics are summarized in Table~\ref{table_1}. Following prior work \cite{piao,yu2020mooccube,bulathwela2021peek}, we train and validate AMR on MOOCCube data from 2017-01-01 to 2019-10-31 and evaluate it on data from 2019-11-01 to 2019-12-31. For PEEK, we use the train-test split provided in the original paper.




\subsection{Evaluation}

We evaluate the top-$k$ KC predictions for learners by computing metrics for every group of 100 KCs in the test set, where one KC interacts with a learner and the remaining 99 does not. For each interacted KC associated with learner $l$, we create the recommendation list $R_l = \{r^1_l, r^2_l, ..., r^k_l\}$, where $r^i_l$ is the KC ranked at the $i$-th position in $R_l$ based on predicted scores. We use two evaluation metrics: \textit{Hit Ratio} of top-$k$ items (HR@K) measures the percentage of actually interacted KCs in the test set that appear within the top-$k$ recommendations, calculated as $HR@k = \frac{1}{N} \sum_{l} I(|R_l \cap T_l|)$, where $N$ is the total number of learners in the test set, $I(x)$ is an indicator function that equals 1 if $x > 0$, and $T_l$ is the set of actually interacted items for learner $l$. \textit{Normalized Discounted Cumulative Gain} of top-$k$ items (nDCG@K) measures ranking quality by accounting for the positions of actually interacted KCs, normalized by the ideal top-$k$ ranking. \textit{nDCG@K} is computed as $\frac{1}{Z}DCG@k = \frac{1}{Z} \sum_{j=1}^{k} \frac{2^{I(|{r^j_l} \cap T_l|)-1}}{\log_2(j+1)}$, where $Z$ is a normalization constant representing the maximum possible \textit{DCG@K} \cite{gong} from an ideal top-$k$ ranking, and $T_l$ is the set of actually interacted items for learner $l$.



\subsection{Baselines}

To compare MOOC recommendation performance, we use seven widely used graph-based baseline models: Metapath2vec \cite{dong2017metapath2vec}, ACKRec \cite{gong}, MOOCIR \cite{piao}, AMCGRec \cite{wang2022attentional}, PGPR \cite{xian2019reinforcement}, CAFE \cite{xian2020cafe}, and UCPR \cite{tai2021user}. PGPR, CAFE, and UCPR, originally proposed for e-commerce recommendation, are included to assess their transferability to MOOC recommendations. All baselines are applicable to our task. For a fair comparison, we reproduce the models following the experimental settings and hyperparameters in their original papers. To implement AMR, the node embedding size is set to 10, the metapath length to 5, and the number of aspects to 5, with training for 50 epochs. Early stopping is applied based on the best \textit{nDCG@5} performance. Training is optimized using the Adam optimizer with a learning rate of $10^{-2}$. 




\section{Results}
\label{result}
\subsection{Main Performance Comparison (RQ1)}

\begin{table}[t]
\centering
\scriptsize
\caption{Recommendation performance on MOOCCube and PEEK measured by different metrics including \textit{HR@K} and \textit{nDCG@K}. The best results are highlighted in \textbf{bold}, and the second-best results are \underline{underlined}.}
\begin{tabular}{|c|c|c|c|c|c|c|c|}
\hline
\textbf{Dataset} & \textbf{Model} & \textbf{HR@5} & \textbf{HR@10} & \textbf{HR@20} & \textbf{nDCG@5} & \textbf{nDCG@10} & \textbf{nDCG@20} \\ \hline
\multirow{8}{*}{\centering MOOCCube} & Metapath2vec & 0.642          & 0.774          & \underline{0.873}    & 0.468          & 0.511          & 0.537          \\
         & ACKRec       & 0.659          & 0.764          & 0.842          & 0.503          & 0.538          & 0.557          \\
         & MOOCIR       & 0.659          & 0.836          & 0.844          & 0.520          & 0.563          & 0.562          \\
         & AMCGRec      & \underline{0.726}    & \underline{0.846}    & 0.851          & \underline{0.541}    & \underline{0.582}    & \underline{0.593}    \\
         & PGPR         & 0.633          & 0.742          & 0.841          & 0.463          & 0.486          & 0.490          \\
         & CAFE         & 0.630          & 0.741          & 0.835          & 0.452          & 0.461          & 0.470          \\
         & UPCR         & 0.604          & 0.733          & 0.821          & 0.450          & 0.480          & 0.487          \\
         & AMR (Ours)   & \textbf{0.754} & \textbf{0.871} & \textbf{0.934} & \textbf{0.581} & \textbf{0.619} & \textbf{0.635} \\ \hline
\multirow{8}{*}{\centering PEEK}     & Metapath2vec & 0.401          & 0.415          & 0.435          & 0.350          & 0.386          & 0.428          \\
         & ACKRec       & 0.432          & 0.452          & 0.494          & 0.352          & 0.391          & 0.419          \\
         & MOOCIR       & 0.480          & 0.520          & 0.516          & \underline{0.451}    & 0.467          & 0.489          \\
         & AMCGRec      & \underline{0.500}    & \underline{0.523}    & \underline{0.539}    & \textbf{0.522} & \underline{0.530}    & \underline{0.534}    \\
         & PGPR         & 0.402          & 0.413          & 0.450          & 0.363          & 0.387          & 0.426          \\
         & CAFE         & 0.428          & 0.432          & 0.442          & 0.350          & 0.373          & 0.421          \\
         & UPCR         & 0.413          & 0.428          & 0.468          & 0.355          & 0.384          & 0.443          \\
         & AMR (Ours)   & \textbf{0.530} & \textbf{0.536} & \textbf{0.579} & \textbf{0.522} & \textbf{0.540} & \textbf{0.551} \\ \hline
\end{tabular}%
\label{table_2}
\end{table}

We evaluate AMR against the previously introduced baseline models to examine overall MOOC recommendation performance. The comparison results, shown in Table \ref{table_2}, reveal that ACKRec, MOOCIR, and AMCGRec, which leverage heterogeneity in learner–KC relationships and rich metapath information, consistently outperform PGPR, CAFE, and UCPR in overall recommendation performance. This suggests that models designed to capture learning patterns and structured relationships in MOOC environments are more effective than general-purpose recommendation systems. Specifically, except for HR@20 in MOOCCube, AMCGRec achieves the second-highest performance on all metrics, indicating that structural neighbor-enriched contrastive learning captures node characteristics more effectively, improving learner preference understanding \cite{wang2022attentional}. Finally, AMR surpasses all baselines on every metric, demonstrating that its ability to consider multiple aspects within paths captures fine-grained learner–KC interactions, delivering accurate and robust recommendations. Additionally, AMR’s random path exploration offers a key advantage over traditional graph-based methods that rely on pre-defined metapaths. Its high performance helps learners stay engaged and complete courses by suggesting content aligned with their interests, enhancing personalized learning and satisfaction. Since both the baseline models and AMR generally achieve higher performance on the MOOCCube dataset than on the PEEK dataset, we conduct the remaining experiments on MOOCCube.

\begin{table}[t]
\centering
\caption{Performance changes on MOOCCube measured by different metrics including \textit{HR@K} and \textit{nDCG@K} across GNN variants of AMR. The best results are highlighted in \textbf{bold}, and the second-best results are \underline{underlined}.}
\label{table_5}
\scriptsize
\begin{tabular}{|c|c|c|c|c|c|c|}
\hline
\textbf{Variants} &   \textbf{HR@5} &  \textbf{HR@10} & \textbf{HR@20} &  \textbf{nDCG@5} &  \textbf{nDCG@10} &  \textbf{nDCG@20} \\
\hline
$AMR_{GCN}$       &         \textbf{0.754} &          \textbf{0.871} &          \textbf{0.934} &    \textbf{0.581} &     \textbf{0.619} &     \textbf{0.635}\\
$AMR_{GAT}$       &         \underline{0.679} &          \underline{0.802} &          \underline{0.891} &    0.532 &     \underline{0.569} &     0.562\\
$AMR_{GraphSAGE}$       &         0.675 &          0.793 &          0.888 &    \underline{0.540} &     \underline{0.569} &     \underline{0.563} \\
\hline
\end{tabular}
\end{table}

\subsection{Performance Comparison of AMR across GNN Variants (RQ1)}


To examine how different GNN backbones impact AMR performance, we implement three variants on the MOOCCube dataset: $AMR_{GCN}$, $AMR_{GAT}$, and $AMR_{GraphSAGE}$. The results in Table \ref{table_5} show that $AMR_{GCN}$ consistently achieves the highest scores across all metrics (HR@5/10/20 and nDCG@5/10/20). This demonstrates that incorporating GCN \cite{kipf} enables more effective neighbor aggregation and richer contextual representations than GAT \cite{velickovic2017graph} or GraphSAGE \cite{hamilton2017inductive}. Lower HR and nDCG scores of $AMR_{GAT}$ and $AMR_{GraphSAGE}$ suggest that attention-based or sample-based aggregation alone cannot capture the fine-grained structural modeling needed for multi-aspect representation. Overall, $AMR_{GCN}$’s superior performance validates the choice of GCN as AMR’s backbone and highlights its key role in improving MOOC recommendation precision.


\begin{figure}[t]
    \centerline{\includegraphics[width=\textwidth]{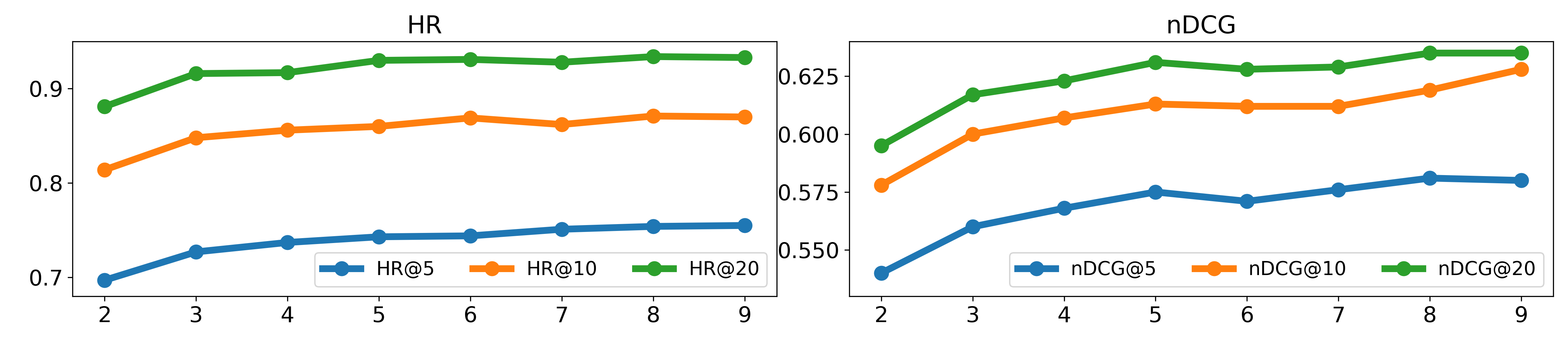}}
    \caption{Performance changes on MOOCCube measured by different metrics including \textit{HR@K} and \textit{nDCG@K} depending on the number of aspects. The performance improves as the number of aspects increases showing the best performance with eight aspects.}
    \label{performance_aspect}
\end{figure}

\subsection{Effect of the Number of Aspects (RQ2)}

To investigate how the number of aspects influences the performance of our method's recommendation, we conduct an ablation study on the MOOCCube dataset, varying the number of aspects from 2 to 9. The performance changes are shown in Fig. \ref{performance_aspect}. The results reveal that model performance improves as the number of aspects increases, suggesting that more aspects enable finer exploration of the diverse attributes in the metapaths, leading to more effective representation of learner-KC relationships. Such fine-grained aspect modeling allows the recommendation system to better capture the heterogeneous characteristics of different learners and KCs in MOOC environments.

\subsection{Influence of Path Length between Learner and KC Nodes (RQ2)}

\begin{figure}[t]
    \centerline{\includegraphics[width=\textwidth]{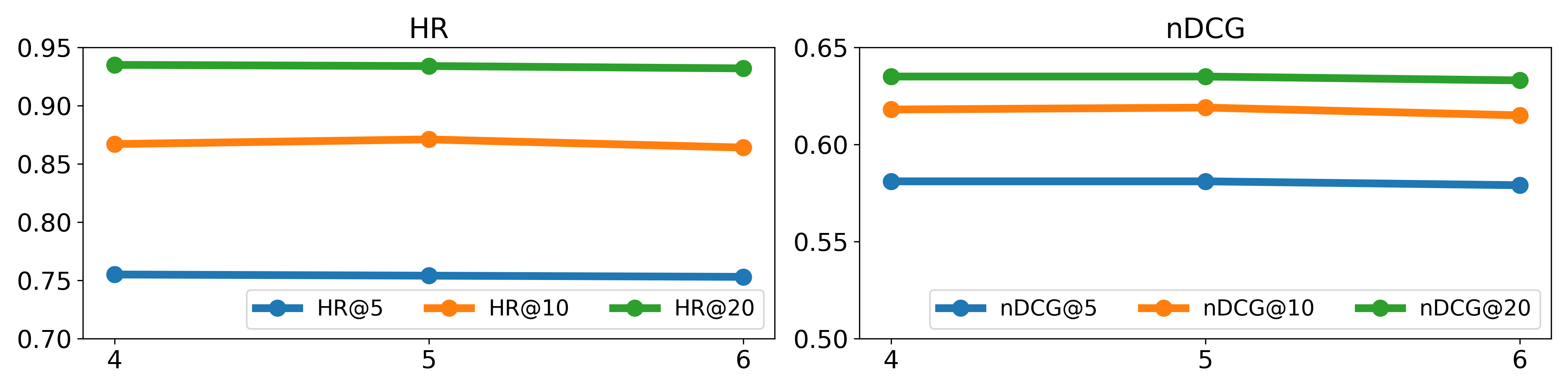}}
    \caption{Performance changes on MOOCCube measured by different metrics including \textit{HR@K} and \textit{nDCG@K} depending on the path length. Model performance remains unchanged as path length varies.}
    \label{performance_path_length}
\end{figure}

To examine the impact of metapath length on recommendation performance, we conduct an ablation study on the MOOCCube dataset, varying the path length to 4, 5, and 6. The results, shown in Fig. \ref{performance_path_length}, indicate that performance is not strongly affected by path length, with the best performance achieved at a path length of 5. This suggests that, in MOOC recommendation, the number of aspects derived from node attributes along a path has a greater influence than the number of inter-node connections. In other words, the quality and richness of aspect-level representations between nodes are more critical than path length in improving recommendation effectiveness.

\subsection{Distribution of Importance Score of Aspects (RQ2)}



\begin{figure}[t]
    \centerline{\includegraphics[width=0.9\textwidth,height=3cm]{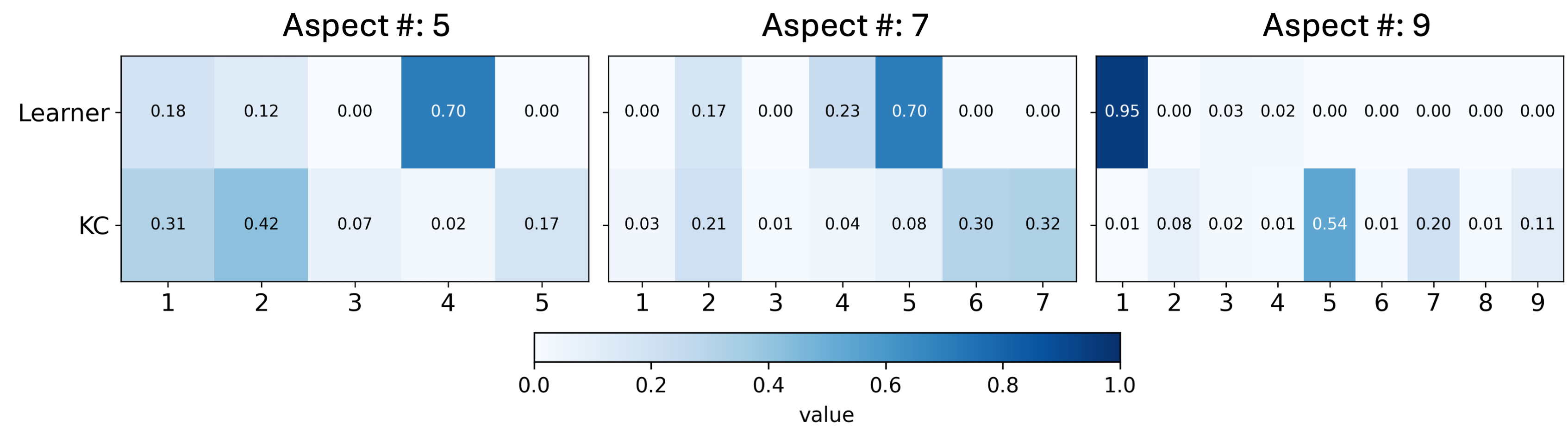}}
    \caption{Distribution of aspect importance for learner and KC representations. KC representations show evenly distributed aspect importance, whereas learner representations are dominated by a few key aspects.}
    \label{aspect_importance}
\end{figure}

Fig. \ref{aspect_importance} shows the distribution of aspect importance for both learner and KC representations on the MOOCCube dataset. A clear contrast emerges: KCs have a relatively even distribution across multiple aspects, indicating the model captures a broad range of attribute combinations. In contrast, learner representations are dominated by two or three aspects. To further explore this, we compute the average frequency of path types discovered by the bi-directional walk: learner-learner paths ($\approx 541.89$) and KC-KC paths ($\approx 3969.06$). KC–KC paths are significantly more numerous, consistent with Table \ref{table_1}, which shows that KC-related relations are more prevalent than learner-related ones. These findings highlight the importance of path diversity for AMR to capture multiple aspects, suggesting the need for future methods to model diverse aspects in networks with fewer paths. This remains an avenue for future research.

\begin{figure}[t]
    \centerline{\includegraphics[width=\textwidth,height=2cm]{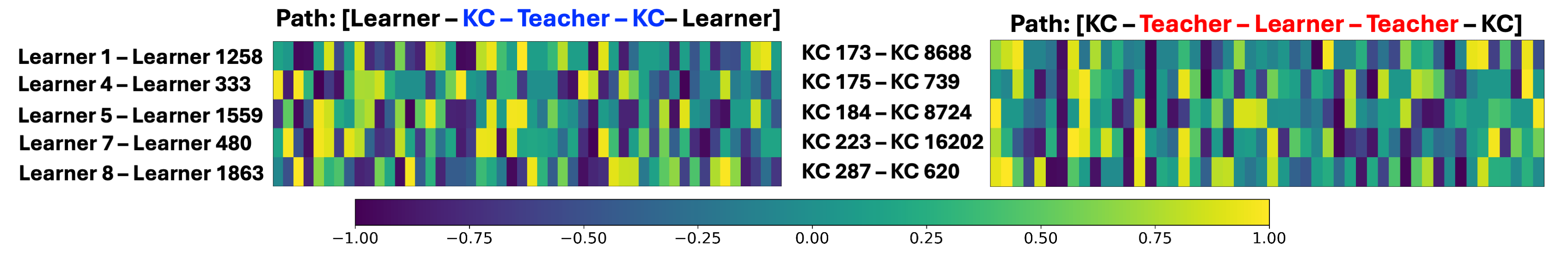}}
    \caption{Diversity of edge features within identical metapaths for learners and KCs. Feature embeddings are 50-dimensional, with colors indicating feature values. The embeddings vary markedly across both learner and KC pairs.}
    \label{aspect_embedding}
\end{figure}

\subsection{Diversity of Edge Features within Identical Metapaths (RQ2)}


AMR extracts diverse aspects from the specific content of nodes in each metapath and uses them as edge features between learners or KCs, producing more effective representations than conventional graph-based methods. To assess edge feature diversity, we analyze their variation among nodes connected through metapaths of the same type on the MOOCCube dataset. The results in Fig. \ref{aspect_embedding} show that edge features between learners or KCs connected by the same metapath (learner: \textcolor{blue}{\textit{KC–teacher–KC}}, KC: \textcolor{red}{\textit{teacher–learner–teacher}}) are highly diverse. This demonstrates that, unlike traditional HIN-based approaches that treat metapaths of the same type as equivalent, AMR captures path-specific representations by emphasizing different aspects according to the nodes within each metapath, enabling a more fine-grained understanding of MOOC-related entity networks.

\section{Discussion, Conclusion, and Future Work}
\label{conclusion}

This paper introduces AMR, a novel framework for recommending KCs in large-scale MOOC environments. AMR overcomes key limitations of traditional metapath-based methods by embedding multiple path-specific aspects derived from node content within each metapath. Extensive experiments on the MOOCCube and PEEK datasets show that AMR consistently outperforms competitive GNN-based baselines across multiple evaluation metrics. Key findings from the analysis of AMR’s relational modeling include: (i) KC representations exhibit balanced aspect distributions, while learner representations are dominated by a few key aspects; (ii) edge features vary substantially within identical metapath types, indicating that AMR’s aspect-based representations go beyond the structural patterns of previous HIN-based approaches. These findings demonstrate that fine-grained relationship representations improve KC recommendation accuracy, enhancing personalized learning and learner satisfaction. However, several limitations remain: (1) path density affects aspect modeling, as KC–KC paths outnumber learner–learner paths; future work should explore methods to model aspects in sparser subgraphs; (2) the fixed number of aspects could be improved with adaptive or data-driven aspect selection. Overall, AMR shows that aspect-aware path representations enhance heterogeneous graph modeling for education, offering a foundation for scalable, personalized learning systems.


%
%
%
\bibliographystyle{splncs04}
\bibliography{reference}





\end{document}